\documentclass[conference]{IEEEtran}
\IEEEoverridecommandlockouts
\usepackage{cite}
\usepackage{amsmath,amssymb,amsfonts}
\usepackage{algorithmic}
\usepackage{graphicx}
\usepackage{textcomp}
\usepackage{xcolor}

\usepackage{graphbox}
\usepackage{multicol}
\usepackage{hyperref}

\def\bg#1{\mbox{\boldmath$#1$}} %for bold greek letters in math mode%

\DeclareMathOperator*{\argmax}{argmax}

\def\BibTeX{{\rm B\kern-.05em{\sc i\kern-.025em b}\kern-.08em
    T\kern-.1667em\lower.7ex\hbox{E}\kern-.125emX}}
\begin{document}

\title{Gated-ViGAT: Efficient Bottom-Up Event Recognition and Explanation Using a New Frame Selection Policy and Gating Mechanism
\thanks{This work was supported by the EU Horizon 2020 programme under grant agreement 101021866 (CRiTERIA).}
}

\author{\IEEEauthorblockN{Nikolaos Gkalelis}
\IEEEauthorblockA{\textit{CERTH-ITI} \\
Thessaloniki, Greece, 57001 \\
gkalelis@iti.gr}
\and
\IEEEauthorblockN{Dimitrios Daskalakis}
\IEEEauthorblockA{\textit{CERTH-ITI} \\
Thessaloniki, Greece, 57001 \\
dimidask@iti.gr}
\and
\IEEEauthorblockN{Vasileios Mezaris}
\IEEEauthorblockA{\textit{CERTH-ITI} \\
Thessaloniki, Greece, 57001 \\
bmezaris@iti.gr}
}

\maketitle

\begin{abstract}
In this paper, Gated-ViGAT, an efficient approach for video event recognition, utilizing bottom-up (object) information, a new frame sampling policy and a gating mechanism is proposed\footnote{Source code is made publicly available at: \url{ https://github.com/bmezaris/Gated-ViGAT}}.
Specifically, the frame sampling policy uses weighted in-degrees (WiDs), derived from the adjacency matrices of graph attention networks (GATs), and a dissimilarity measure to select the most salient and at the same time diverse frames representing the event in the video.
Additionally, the proposed gating mechanism fetches the selected frames sequentially, and commits early-exiting when an adequately confident decision is achieved.
In this way, only a few frames are processed by the computationally expensive branch of our network that is responsible for the bottom-up information extraction.
The experimental evaluation on two large, publicly available video datasets (MiniKinetics, ActivityNet) demonstrates that Gated-ViGAT provides a large computational complexity reduction in comparison to our previous approach (ViGAT), while maintaining the excellent event recognition and explainability performance.\footnote{\copyright2022 IEEE. Personal use of this material is permitted. Permission from IEEE must be obtained for all other uses, in any current or future media, including reprinting/republishing this material for advertising or promotional purposes, creating new collective works, for resale or redistribution to servers or lists, or reuse of any copyrighted component of this work in other works.}
\end{abstract}


\section{Introduction}
\label{sec:intro}

\IEEEPARstart{O}{ver} the last years an increasing number of real-world applications in various sectors, such as multimedia \cite{GhodratiCVPR2021} and medicine \cite{Esteva2021}, to name a few, resort to automated event recognition techniques in order to increase the quality of the provided services.
Deep learning techniques have achieved major performance leaps and new improvements in this domain continue to push the recognition performance limits every year \cite{GhodratiCVPR2021,GaoNIPS21,ZhouTMM2022}. These methods usually operate in a top-down fashion, i.e., a neural network is trained using the video class labels and entire frames (or video segments) to implicitly learn to focus on the video regions that are mostly related with the occurring event.

Studies in cognitive science have suggested that humans interpret complex scenes by selecting a subset of the available sensory information in a bottom-up manner, most probably in order to reduce the complexity of scene analysis \cite{TsotsosArtifIntel1995,IttiPAMI1998,Chi_CVPR_2020}.
To this end, bottom-up event recognition approaches, which support the classifier by providing the main objects depicted in the frames, are recently getting increasing attention \cite{GkalelisCVPR2021,Gkalelis2022}.
These methods not only improve the recognition accuracy but also provide object- and frame-level explanations about the classifier's outcome. 
However, due to the extraction and processing of the bottom-up (object) information, these methods have a quite high computational cost at inference time, restricting their applicability in applications with strict latency constraints.

\begin{figure}[!t]
\centering
\includegraphics[width=0.99\columnwidth]{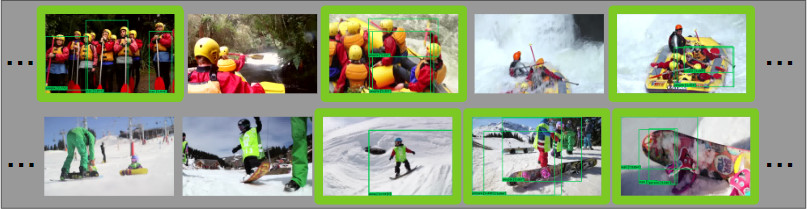}
\caption{Gated-ViGAT in action.
Frames of two test videos from ActivityNet \cite{caba2015activitynet} are shown, belonging to the events ``Rafting'' (top row) and ``Snowboarding'' (bottom row).
Gated-ViGAT classifies correctly both videos using only three frames (i.e. the ones shown within a green rectangle), derived using our new frame selection policy.
The selected frames are relevant to the event and at the same diverse, providing a broad view of the video event, and, thus, a better input for the categorization; e.g., in the bottom row, we can see different skiers and a snowboard.
Moreover, bottom-up (object) information provides additional cues for understanding why a video is correctly classified or not.
}
\label{fig:ConceptualOverview}
\end{figure}
Inspired from recent efficient top-down approaches \cite{GhodratiCVPR2021}, which try to reduce the computational cost of high-capacity 2D or 3D convolutional neural networks (CNNs), here, we extend our previously proposed method, ViGAT \cite{Gkalelis2022}, using a new frame selection policy and a gating mechanism.
Thus, in contrast to ViGAT, the proposed approach, called hereafter Gated-ViGAT, extracts bottom-up information from only a small fraction of the sampled frames as shown in Fig. \ref{fig:ConceptualOverview}.
The proposed frame selection policy utilizes an explanation and a dissimilarity measure (similarly to works in other domains, e.g. \cite{apostolidis_ICMR_22}) to select the frames that better represent the event depicted in the video as well as provide a diverse overview of it.
Additionally, we replace the CNN-based gating mechanism of \cite {GhodratiCVPR2021} with one that combines both convolution and attention \cite{Gkalelis2022,WuIccv2021} in order to be able to process sequences of frames (not only individual frames as in \cite {GhodratiCVPR2021}) and thus capture more effectively both the short- and long-term dependencies of the event occurring in the video.
Consequently, the proposed Gated-ViGAT, continues to achieve a high recognition performance, as ViGAT, but with a significant computational complexity reduction.
Moreover, contrarily to efficient top-down approaches, e.g. \cite{Meng20,GhodratiCVPR2021}, it can provide explanations about the classification outcome, as we demonstrate with a comprehensive qualitative study.  
We evaluate the proposed method in two large, publicly available video datasets (MiniKinetics \cite{XieECCV18}, ActivityNet \cite{caba2015activitynet}), verifying its efficacy.
In summary, our major contributions are:
\begin{itemize}
\item We present a new frame selection policy and a gating mechanism, and adapt them to our recently proposed bottom-up approach achieving a considerable computational complexity reduction at inference stage.

\item The proposed Gated-ViGAT retains the high recognition performance of ViGAT, outperforming the best top-down approaches, and in comparison to them, can provide object- and frame-level explanations about the event recognition outcome.
\end{itemize}

\section{Related Work}
\label{sec:RelatedWork}

We survey video recognition approaches that are mostly related to ours.
For a broader literature survey the interested reader is referred to \cite{ZhuArxiv2020,PareekASO2021}.

\subsection{Event and action recognition}

A major focus on this area is the design of approaches that capture more effectively the long-term dependencies of events/actions in videos. 
For instance, in \cite{Kaiyu_NIPS_2018}, a non-local module modifies a 2D ResNet backbone in order to better capture the action dynamics in videos.
Similarly, in \cite{GaoNIPS21}, an attentive pooling mechanism is applied in various CNN networks to combine frame-level action recognition scores.
In \cite{FayyazCVPR21}, a mechanism inserted in 3D-CNNs adaptively adjusts the temporal resolution of feature maps depending on the complexity of the action.
In \cite{ZhouTMM2022}, a local- and a global-branch are used to extract semantic and temporal action information, respectively.
Video vision transformer (ViViT) \cite{Arnab_2021_ICCV} factorizes attention to spatial and temporal dimensions in order to effectively process long video sequences.
Differently from the above approaches, ViGAT \cite{Gkalelis2022} uses an object detector to extract bottom-up (i.e. object) information from video frames, a Vision Transformer (ViT) backbone to derive a feature representation for each object and frame, and an attention-based head network to recognize and explain events in video.
The above methods have considerably improved the event/action recognition performance, however, employ networks with quite high computational complexity imposing limitations to their widespread applicability.

\subsection{Frame selection policies}

The utilization of frame selection policies for decreasing the computational complexity of event/action recognition approaches is an area currently receiving increasing attention \cite{WuCVPR19,WuICCV19,GaoCVPR2020,Meng20,GowdaAAAI21,GhodratiCVPR2021}.
AdaFrame \cite{WuCVPR19} exploits a policy gradient method to select future frames for efficient video recognition.
SCSampler \cite{KorbarICCV19} utilizes the audio modality and an agent to discard redundant video clips.
Similarly, an audio-based previewing tool is used by ListenToLook \cite{GaoCVPR2020} to select the most salient video frames.
In \cite{WuICCV19}, the frame sampling policy is formulated as multiple parallel Markov decision processes and learned using multi-agent reinforcement learning (MARL).
SMART \cite{GowdaAAAI21} selects the most discriminant frames using a multi-frame attention and relation network.
FrameExit \cite{GhodratiCVPR2021} combines a deterministic frame sampling policy with conditional exiting, i.e., frames are processed until a sufficiently confident decision is reached.
The above policies have successfully applied to several top-down video recognition approaches; however, the applicability of such policies to methods that utilize bottom-up video information to the best of our knowledge is an unexplored topic.

\begin{figure}[!ht]
\centering
\includegraphics[width=\columnwidth]{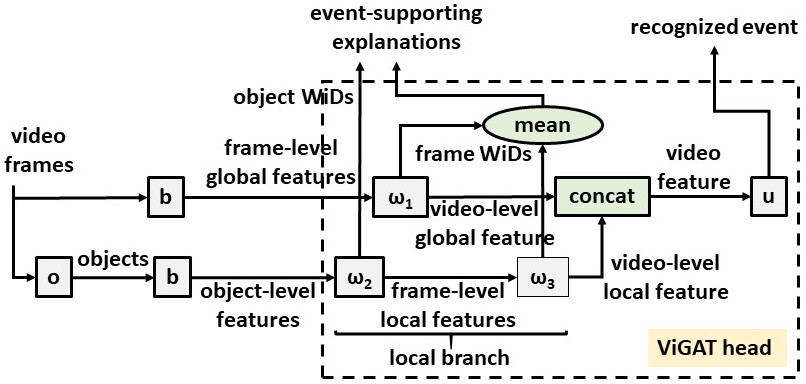}
\caption{ViGAT's block diagram \cite{Gkalelis2022}.
ViGAT consists of an object detector $o$, a network backbone $b$ and the ViGAT head.
The latter consists of three GAT blocks, $\omega_1$, $\omega_2$, $\omega_3$, that process the global (frame-level) and the local (object-level) features, and a dense layer $u$ that uses the video-level feature to recognize the event occurring in the video.
Moreover, using the weighted in-degrees (WiDs) of GAT blocks' adjacency matrices, ViGAT can provide event-supporting explanations at object- and frame-level.
Gated-ViGAT utilizes the pretrained components of ViGAT and introduces a gating component to build a new architecture for efficient event recognition and explanation.}
\label{fig:ViGatBlockDiagram}
\end{figure}

\section{Proposed method}
\label{sec:ProposedMethod}

\begin{figure*}[!ht]
\centering
\includegraphics[width=1.69\columnwidth]{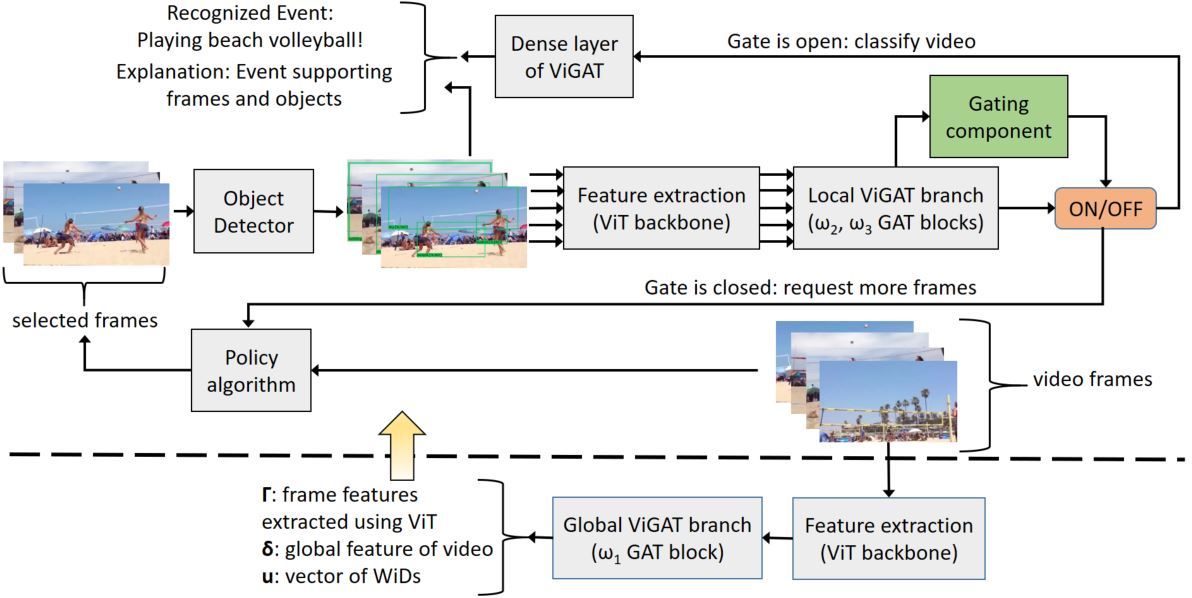}
\caption{Illustration of the proposed Gated-ViGAT.
Our architecture utilizes the pretrained components of ViGAT, and the proposed frame selection policy and gating component.
The policy algorithm requests sequentially an increasing number of frames until the gating component commits an exiting action, signaling that the provided frames are adequate for recognizing the underlying video event (here ``Playing beach voleyball'').}
\label{fig:GatedViGAT}
\end{figure*}

Gated-ViGAT uses the pretrained components of ViGAT \cite{Gkalelis2022}, whose block diagram is shown in Fig. \ref{fig:ViGatBlockDiagram}.
Specifically, Gated-ViGAT utilises the object detector $o$, the Vision Transformer (ViT) backbone $b$ \cite{DosovitskiyICLR21},
and the three GAT blocks \cite{Velickovic_ICLR_2018} and dense layer of the ViGAT head, denoted as $\omega_1$, $\omega_2$, $\omega_3$ and $u$, respectively.
In addition to the above, Gated-ViGAT introduces a gating component \cite{GhodratiCVPR2021} consisting of $S$ gates, $g^{(1)}, \dots, g^{(S)}$, whose role is to identify the minimum number of frames to process in the local feature branch ($\omega_2$, $\omega_3$) of the ViGAT, so that the classification performance is retained at the highest degree.

The gate networks $g^{(s)}$ are identical, composed of a one-dimensional CNN, a GAT block (as the ones used in ViGAT \cite{Gkalelis2022}), and a dense layer that outputs a value in the range [0,1],
\begin{equation}
    g^{(s)}: \mathbb{R}^{s+1 \times F} \rightarrow [0,1], \; s=1, \dots, S, \label{E:Gates}
\end{equation}
where, $F$ is the feature vector dimensionality produced by the ViT backbone.
The $s$th gate is used to decide whether $Q^{(s)}$ frames are enough to yield a confident decision for the input video.
In case that the output of the $s$th gate is smaller than 0.5 (i.e. the gate is closed), additional frames are selected, and fetched by the local feature branch of ViGAT and the next gate, $g^{(s+1)}$.
This process continues until a gate opens (i.e. its output is larger than 0.5) or when the last gate is reached.

The $Q^{(s)}$ frames for the $s$th gate are selected using a novel frame selection policy (Section \ref{sss:Policy}) that utilizes the most salient frame (in terms of explanation) and a dissimilarity criterion in order to achieve high coverage of the event occurring in the video.
The gating component is trained end-to-end, while the rest of Gated-ViGAT's components are  kept frozen, as explained in the following. This allows us to achieve comparable results while dramatically reducing the computational complexity. 

\subsection{Gate training}
\label{ss:GateTrain}

Suppose an annotated video dataset of $N$ videos and $G$ classes.
Each video $\mathcal{V}$ is represented with
$P$ frames,
\begin{equation}
    \mathcal{V} = \{\mathbf{V}_p\}_{p=1}^{P}, \label{E:vidframes}
\end{equation}
and is associated with a class vector $\mathbf{y} = [y_1, \dots, y_G]^T \in \{0,1\}^{G}$,
where, $\mathbf{V}_p \in \mathbb{R}^{w \times h \times c}$ is the $p$th frame, $w$, $h$ and $c$ are its width, height and number of channels, and $y_g$ is one if the video $\mathcal{V}$ belongs to class $g$ and zero otherwise.

\subsubsection{Global feature vectors}
\label{sss:GlobalFea}

A ViT backbone $b$ is used to represent each frame with a global feature vector $\mathbf{\gamma}_p \in \mathbb{R}^F$,
\begin{equation}
    \bg{\gamma}_p = b(\mathbf{V}_p). \label{E:globalFrameFea}
\end{equation}
The sequence of feature vectors corresponding to the video are then stacked row-wise forming a matrix $\mathbf{\Gamma} = [\bg{\gamma}_1, \dots, \bg{\gamma}_P]^T$ and passed to the block $\omega_1$ of the trained ViGAT head to derive a new feature representation $\bg{\delta} \in \mathbb{R}^F$ of the video along with a vector of WiDs $\mathbf{u} \in [0,1]^P$,
\begin{equation}
    \bg{\delta}, \mathbf{u} = \omega_1(\mathbf{\Gamma}). \label{E:globalVidFea}
\end{equation}
Note that the $p$th element of $\mathbf{u}$ corresponds to 
the respective video frame and can be used to estimate its importance for explaining model's decision \cite{Gkalelis2022}.

\subsubsection{Frame selection policy}
\label{sss:Policy}

The following algorithm is used to select $Q^{(s)}$ frames for the first gate (i.e. $s$ is initially set to 1).
Firstly, we compute the index $p_1$ of the first frame using
\begin{equation}
     p_1 = \argmax (\mathbf{u}), \label{E:policyFirstFrame}
\end{equation}
i.e., the first frame is the one that corresponds to the highest WiD value, and normalize all the global feature vectors using
\begin{equation}
      \tilde{\bg{\gamma}}_p = \frac{\bg{\gamma}_p}{\| \bg{\gamma}_p \|}, \label{E:normOneGlobalFeaVec}
\end{equation}
where $\| \bg{\gamma} \|$ is the $L_2$ norm of any vector $\bg{\gamma}$.
A dissimilarity score $\alpha_p$ of the selected frame with all other frames of the video is then computed using
\begin{equation}
    \alpha_p = \frac{1}{2} (1 - \tilde{\bg{\gamma}}_{p_1}^T \tilde{\bg{\gamma}}_{p}).   \label{E:frameDissimilarity}
\end{equation}
Note that $\alpha_p \in [0,1]$ because $\tilde{\bg{\gamma}}_{p_1}^T \tilde{\bg{\gamma}}_{p} \in [-1, 1]$ as it is the cosine distance between the vectors $\tilde{\bg{\gamma}}_{p_1}$ and $\tilde{\bg{\gamma}}_{p}$.
Next, both the dissimilarity scores $\alpha_p$ and the WiD values in $\mathbf{u}$ are scaled in the range $[0,1]$ using min-max normalization,
\begin{eqnarray}
    \tilde{\alpha}_p &=& \frac{\alpha_p - \alpha_{min}}{\alpha_{max} - \alpha_{min}}, \label{E:disimMinMax} \\
    \tilde{u}_p &=& \frac{u_p - u_{min}}{u_{max} - u_{min}}, \label{E:widMinMax}
\end{eqnarray}
where, $\alpha_{max} = \max(\mathbf{\bg{\alpha}})$, $\alpha_{min} = \min(\mathbf{\bg{\alpha}})$, $u_{max} = \max(\mathbf{u})$, $u_{min} = \min(\mathbf{u})$ and $\bg{\alpha} = [\alpha_1, \dots, \alpha_P]^T$.
The dissimilarity scores are then multiplied with the respective normalized WiDs to update the vector of WiDs,
\begin{equation}
    u_p = \tilde{u}_p \odot \tilde{\alpha}_p, \label{E:widUpdate}
\end{equation}
where $\odot$ denotes element-wise multiplication.
The same procedure (Eqs. (\ref{E:policyFirstFrame}) to (\ref{E:widUpdate})) is followed to select the rest $Q^{(s)}-1$ frames for the $s$th gate.
Along the different gates, the frames are added incrementally, i.e., for gate $s+1$, the $Q^{(s)}$ frames selected for gate $s$ are retained and the same procedure (Eqs. (\ref{E:policyFirstFrame}) to (\ref{E:widUpdate})) is followed to select the $Q^{(s+1)} - Q^{(s)}$ additional frames for gate $s+1$.
The number of frames per gate, $Q^{(1)}, \dots, Q^{(S)}$, is a design parameter defined by the user.

\subsubsection{Local feature vectors}
\label{sss:LocalFea}

Suppose $\{ \mathbf{V}_{p_\iota} \}_{\iota=1}^{Q^{(s)}}$ are the frames selected for gate $s$.
The ViGAT object detector is used to derive $K$ objects from each selected frame, represented with a bounding box, an object class label and the associated degree of confidence (DoC).
The ViT backbone is applied to represent each object with a feature vector; these feature representations are sorted in descending order using their DoC values and stacked column-wise to form a matrix $\mathbf{X}_{p_\iota} \in \mathbb{R}^{K \times F}$ for the $p_\iota$th selected frame,
\begin{equation}
    \mathbf{X}_{p_\iota} = [\mathbf{x}_{p_\iota,1}, \dots, \mathbf{x}_{p_\iota,K}]^T. \label{E:objectVitFea}
\end{equation}
The above matrix is fetched by the block $\omega_2$ to derive a ``local'' feature vector $\bg{\eta}_{p_\iota} \in \mathbb{R}^{F}$ and a vector of WiDs $\bg{\phi}_{p_\iota} \in [0,1]^K$,
\begin{equation}
    \bg{\eta}_{p_\iota}, \bg{\phi}_{p_\iota} = \omega_2(\mathbf{X}_{p_\iota}). \label{E:localFrameFea}
\end{equation}
The $k$th element of $\bg{\phi}_{p_\iota}$ can be used as an indicator for the contribution of the $k$th object in associating the $p_\iota$th frame with the recognized event \cite{GkalelisCVPR2021,Gkalelis2022}.
The above features are used to form a matrix $\mathbf{H}^{(s)} = [\bg{\eta}_{p_1}, \dots, \bg{\eta}_{p_{Q^{(s)}}}]^T \in \mathbb{R}^{Q^{(s)} \times F}$ and fetched by the block $\omega_3$ to derive a ``local'' feature vector $\bg{\varrho}^{(s)} \in \mathbb{R}^{F}$ for the entire video and with respect to gate $s$,
\begin{equation}
    \bg{\varrho}^{(s)} = \omega_3(\mathbf{H^{(s)}}). \label{E:localVidFea}
\end{equation}

\subsubsection{Gate pseudolabels}
\label{sss:GatePseudolabels}

The video-level global (\ref{E:globalVidFea}) and local (\ref{E:localVidFea}) feature vectors are concatenated to form one vector $\bg{\zeta}^{(s)} \in \mathbb{R}^{2F}$ for the entire video with respect to gate $s$,
\begin{equation}
    \bg{\zeta}^{(s)} = [\bg{\delta}; \bg{\varrho}^{(s)}]. \label{E:vidFea}
\end{equation}
This feature vector is fetched by the dense layer $u$ of the trained ViGAT to derive a vector $\hat{\mathbf{y}} = [\hat{y}_1, \dots, \hat{y}_G]^T$, where $\hat{y}_g \in [0,1]$ is the degree of confidence that the video belongs to class $g$.
The standard cross-entropy or the categorical cross-entropy loss for single- or multilabeled datasets, respectively, is used to compute the loss $l$ for the specified video.
A pseudolabel $o^{(s)} \in \{0,1\}$ for video with respect to gate $s$ is then derived using \cite{GhodratiCVPR2021}
\begin{equation}
    o^{(s)} = \left\{\begin{array}{rll}
         1 & \mbox{if} & l \le \epsilon^{(s)}, \\
         0 & \mbox{else} & 
    \end{array}\right. \label{E:gatePseudolabels}
\end{equation}
where, $\epsilon^{(s)}$ determines the minimum loss required to exit gate $s$ (and thus the entire gating component), defined as, $\epsilon^{(s)} = \beta \exp{\frac{s}{2}}$, and a $\beta$ is scalar parameter.

\subsubsection{Gate loss}
\label{sss:Gateloss}

The $S$ gates are trained using the corresponding gate pseudolabels (\ref{E:gatePseudolabels}) and the binary cross-entropy losses, $l_{bce} ()$, summed along all gates \cite{GhodratiCVPR2021},
\begin{equation}
    \mathcal{L} = \frac{1}{S} \sum_{s=1}^S l_{bce} ( g^{(s)}(\mathbf{Z}^{(s)}), o^{(s)})
\end{equation}
where, $\mathbf{Z}^{(s)} \in \mathbb{R}^{s+1 \times F}$, $\mathbf{Z}^{(1)} = [\bg{\delta}, \bg{\varrho}^{(1)}]^T$ and $\mathbf{Z}^{(s)} = [\bg{\delta}, \bg{\varrho}^{(1)}, \dots,  \bg{\varrho}^{(s)}]^T$ for $s>1$.

\subsection{Event recognition and explanation}
\label{ss:Inference}

During the inference stage, the matrix $\acute{\mathbf{Z}}^{(s)}$ for the test video $\acute{\mathcal{V}}$ is generated for $s = 1, \dots, S$, sequentially and each time the output of the respective gate is inquired to decide weather to exit the gating component, or, request the generation of additional local feature vectors.
Suppose that $s^{\ast}$ is the gate for which the video exits the gating component, i.e., $g^{(s^{\ast})} (\acute{\mathbf{Z}}^{(s^{\ast})}) > 0.5$.
At this event, the derived global (\ref{E:globalVidFea}) and local (\ref{E:localVidFea}) feature vectors are concatenated to form a feature vector $\acute{\bg{\zeta}}^{(s^{\ast})} = [\acute{\bg{\delta}}; \acute{\bg{\varrho}}^{(s^{\ast})}]$ for the entire test video (\ref{E:vidFea}), which is fetched by the dense layer $u$ of the trained ViGAT to classify the video to one of the $G$ classes.
Moreover, as explained in previous works \cite{GkalelisCVPR2021,Gkalelis2022} the derived WiDs can be used to provide explanations about the model's event recognition outcome.
To this end, explanations at frame-level are produced using the top frames selected by the proposed frame selection policy (Section \ref{sss:Policy}).
On the other hand, the vector of WiDs $\acute{\bg{\phi}}$ (\ref{E:localFrameFea}) is exploited to derive  explanations at object-level, e.g., by selecting the objects corresponding to the WiDs with highest values \cite{GkalelisCVPR2021}. 

\section{Experiments}
\label{sec:Experiments}

\subsection{Datasets}
\label{ssec:Datasets}

We run experiments on two large, publicly available video datasets:
i) MiniKinetics \cite{XieECCV18} is a subset of the Kinetics dataset \cite{Carreira_2017_CVPR}, consisting of 200 action classes, 80K training and 5K testing video clips.
Each clip is sampled from a different YouTube video, has 10 seconds duration and is annotated  with a single event/action class label.
ii) ActivityNet v1.3 \cite{caba2015activitynet} is a popular multilabel video benchmark consisting of 200 event/action classes (including a large number of high-level events), and approximately 10K, 5K and 5K videos for training, validation and testing, respectively.
Most videos are 5 to 10 minutes long.
As the testing-set labels are not publicly available, the evaluation is performed on the so called validation set, as typically done in the literature.

\subsection{Setup}
\label{ssec:SetupEval}
Uniform sampling is first applied to represent each video with a sequence of $N = 30$ frames for MiniKinetics (e.g. as in  \cite{KimCVPR20,ZhouTMM2022,Gkalelis2022}) and $N = 120$ frames for ActivityNet (e.g. as in  \cite{WuICCV19,Gkalelis2022}).
As explained in Section \ref{sec:ProposedMethod}, the following components are utilized from ViGAT \cite{Gkalelis2022}: a) the Faster R-CNN \cite{renNips2015faster} object detector $o$ pretrained on ImageNet1K and finetuned on Visual Genome dataset, b) the ViT-B/16 backbone $b$ pretrained on ImageNet11K and fine-tuned on ImageNet1K, c) the three GAT blocks, $\omega_1$, $\omega_2$, $\omega_3$, pretrained on MiniKinetics or ActivityNet, depending on the dataset used in the experimental evaluation.
As in ViGAT, the number of objects $K$ (\ref{E:objectVitFea}) to extract with the object detector $o$ is set to 50.

The number of gates $S$ of the gating component and the length of frame sequences $Q^{(s)}$ corresponding to the different gates are set to $S=5$, $\{ Q^{(s)} \}_{s=1}^{5} = \{2, 4, 6, 8, 10\}$ and $S=6$, $\{ Q^{(s)} \}_{s=1}^{6} = \{9, 12, 16, 20, 25, 30\}$ for the MiniKinetics and ActivityNet experiment, respectively.
We used one more gate and larger frame sequences on ActivityNet because it contains arbitrarily large videos in contrast to  MiniKinetics where all videos are rather short (10 secs duration).
In both datasets, the gating component is trained for 40 epochs using an initial learning rate of $10^{-4}$ multiplied by 0.1 at epochs 16 and 35.
Similarly to other works in the literature, the recognition performance is measured using the mean average precision (mAP) and top-1 accuracy on ActivityNet and MiniKinetics, respectively.
All experiments were run on PCs with an Intel i5 CPU and a single RTX3090 NVIDIA GPU.

\begin{table}[!ht]
\caption{Performance comparison on MiniKinetics \cite{XieECCV18} ($^\star$Note that FrameExit is evaluated on the MiniKinetics variant of \cite{Meng20})}
\begin{center}
{%\footnotesize
\begin{tabular}{lc}
       & top-1(\%) \\
\hline
    TBN \cite{LiAAAI19} & 69.5  \\
    BAT \cite{Chi_CVPR_2020} & 70.6 \\
    MARS (3D ResNet backbone)  \cite{CrastoCVPR19} & 72.8  \\
    Fast-S3D (Inception backbone)  \cite{XieECCV18} & 78.0 \\
    ATFR (X3D-S backbone)\cite{FayyazCVPR21} & 78.0\\
    ATFR (R(2+1)D backbone) \cite{FayyazCVPR21} & 78.2 \\
    RMS (SlowOnly backbone) \cite{KimCVPR20} & 78.6 \\
    ATFR (I3D backbone) \cite{FayyazCVPR21}&  78.8 \\
    Ada3D  (I3D backbone on Kinetics) \cite{LiCVPR21}& 79.2 \\
    ATFR (3D Resnet backbone) \cite{FayyazCVPR21} &  79.3 \\
    CGNL (Modified ResNet backbone) \cite{Kaiyu_NIPS_2018} & 79.5 \\
    TCPNet (ResNet backbone on Kinetics) \cite{GaoNIPS21} &  80.7 \\
    LgNet (R3D Backbone) \cite{ZhouTMM2022}& 80.9 \\
    FrameExit  (EfficientNet backbone) \cite{GhodratiCVPR2021}$^\star$ & 75.3 \\ 
    ViGAT \cite{Gkalelis2022} & \textbf{82.1} \\
    Gated-ViGAT (proposed) & \textit{81.3} \vspace{3pt}
    \\
  \end{tabular}}
\end{center}
\label{tbl:ExpResMiniKinetics}
\end{table}

\begin{table}[!ht]
\caption{Performance comparison on ActivityNet \cite{caba2015activitynet}.}
\begin{center}
{%\footnotesize
\begin{tabular}{lc}
       & mAP(\%) \\
\hline
  AdaFrame \cite{WuCVPR19} & 71.5 \\
  ListenToLook \cite{GaoCVPR2020} & 72.3 \\
  LiteEval \cite{WuNIPS19} & 72.7 \\
  SCSampler \cite{KorbarICCV19} & 72.9 \\
  AR-Net \cite{Meng20} & 73.8 \\
  FrameExit \cite{GhodratiCVPR2021} & 77.3 \\ 
  AR-Net (EfficientNet backbone) \cite{Meng20} & 79.7 \\
  MARL (ResNet backbone on Kinetics) \cite{WuICCV19} & 82.9 \\
  FrameExit (X3D-S backbone) \cite{GhodratiCVPR2021} & 87.4 \\
  ViGAT \cite{Gkalelis2022} & \textbf{88.1} \\
  Gated-ViGAT (proposed) & \textit{87.5} 
  \end{tabular}}
\end{center}
\label{tbl:ExpResActnet}
\end{table}

\subsection{Event recognition results}
\label{ssec:EventRecognitResults}

We compare the proposed Gated-ViGAT against the best performing approaches in the literature on the two datasets, namely, TBN \cite{LiAAAI19}, BAT \cite{Chi_CVPR_2020}, MARS \cite{CrastoCVPR19}, Fast-S3D \cite{XieECCV18}, ATFR \cite{FayyazCVPR21}, RMS \cite{KimCVPR20}, Ada3D \cite{LiCVPR21}, CGNL \cite{Kaiyu_NIPS_2018}, TCPNet \cite{GaoNIPS21}, LgNet \cite{ZhouTMM2022}, FrameExit \cite{GhodratiCVPR2021}, AdaFrame \cite{WuCVPR19}, ListenToLook \cite{GaoCVPR2020}, LiteEval \cite{WuNIPS19}, SCSampler \cite{KorbarICCV19}, AR-Net \cite{Meng20}, MARL \cite{WuICCV19} and ViGAT \cite{Gkalelis2022}.
The recognition performance of the various methods on MiniKinetics and ActivityNet are shown in Tables \ref{tbl:ExpResMiniKinetics} and \ref{tbl:ExpResActnet}, respectively.
Most of these methods utilize a ResNet type backbone pretrained on ImageNet; when this is not the case we provide the name of the used backbone in brackets.
We also denote the best and second best recognition rate with bold and italic fonts, respectively.
From these results we see that Gated-ViGAT outperforms all previous methods (which are all top-down) except ViGAT.
This confirms that Gated-ViGAT in comparison to the top-down approaches, utilizes effectively the complementary discriminant event information from the bottom-up features, achieving a performance gain; and additionally, can provide comprehensive explanations as we show in the next subsection.

\begin{table}[!t]
\caption{Performance comparison in terms of computational complexity (TFLOPS) between ViGAT and Gated-ViGAT on two datasets.}
\begin{center}
{%\footnotesize 
\begin{tabular}{l|cc}
     & ViGAT & Gated-ViGAT \\
\hline
    MiniKinetics &  34.4  & \textbf{8.7} \\
    ActivityNet &  137.4  & \textbf{24.8}
  \end{tabular}}
\end{center}
\label{tbl:ComputComplexity}
\end{table}

As mentioned above, ViGAT slightly outperforms the proposed Gated-ViGAT, by 0.8\% mAP and 0.6\% top-1 accuracy in MiniKinetics and ActivityNet, respectively.
This is expected, as in contrast to ViGAT that extracts bottom-up information from every sampled frame, Gated-ViGAT extracts bottom-up information from only a small fraction of them.
However, due to this fact, Gated-ViGAT is much more efficient than ViGAT at inference time.
To quantify this, we used the the Fvcore Flop Counter \cite{fvcorFlopCount} to measure the computational complexity during inference of the above methods in terms of FLOPs (floating point operations) on both MiniKinetics and ActivityNet.
From the obtained results, shown in Table \ref{tbl:ComputComplexity}, we see that Gated-ViGAT provides a dramatic reduction in TFLOPs over ViGAT, specifically, of approximately $\times 4$ and $\times 5.5$ reduction, in MiniKinetics and ActivityNet, respectively.
This is because, the most heavy components of both architectures, i.e., the Faster R-CNN object detector and ViT backbone, which are used to extract and process the bottom-up information (i.e. the $K=50$ objects per frame), are applied considerably fewer times with Gated-ViGAT.
Here, we should also note that as expected, during inference, Gated-ViGAT has a higher computational complexity than most top-down approaches (e.g. \cite{ZhouTMM2022, GaoNIPS21} 
report requiring in the order of tens or hundreds of GFLOPs at inference).
However, as we see in Tables \ref{tbl:ExpResMiniKinetics}, \ref{tbl:ExpResActnet}, Gated-ViGAT has outperformed all top-down approaches evaluated here, and, additionally, in comparison to these approaches, can provide explanations 
concerning the classifier's decision, as shown in Section \ref{ssec:ExplanEventRecognitResults}.

In order to get additional insight of the proposed method, we provide the number of processed test videos, the number of frames used to extract the bottom-up information and the recognition performance, with respect to the individual gates of Gated-ViGAT, for MiniKinetics and ActivityNet, in Tables \ref{tbl:perfGatesMiniKinetics} and \ref{tbl:perfGatesActivityNet}, respectively.
Using this information, the average number of frames per video is estimated to 7 and 20 for MiniKinetics and ActivityNet, respectively.
This is expected as most videos of ActivityNet are longer than the ones of MiniKinetics (see Section \ref{ssec:Datasets}).
We also observe a recognition rate drop as the gate number $s$ increases.
This is due to the fact that the most difficult to recognize videos do not exit the gating component early, thus reducing the performance of the gates towards the end of the component; this is in agreement with similar studies in the literature \cite{GhodratiCVPR2021}.

\begin{table}[!t]
\caption{Number of processed videos and recognition performance (in terms of top-1(\%)) with respect to each gate of Gated-ViGAT on MiniKinetics.}
\begin{center}
{%\footnotesize
\begin{tabular}{l|ccccc}
     & $g^{(1)}$ & $g^{(2)}$ & $g^{(3)}$ & $g^{(4)}$ & $g^{(5)}$ \\
\hline
    \# frames & 2 & 4 & 6 & 8 & 10 \\
    \# videos & 179 & 686 & 1199 & 458 & 2477 \\
    top-1(\%) & 84.9\% & 83\% & 81.1\% & 84.9\% & 80.7\% 
  \end{tabular}}
\end{center}
\label{tbl:perfGatesMiniKinetics}
\end{table}

\begin{table}[!t]
\caption{Number of processed videos and recognition performance (in terms of mAP(\%)) with respect to each gate of Gated-ViGAT on ActivityNet.}
\begin{center}
{%\footnotesize
\begin{tabular}{l|cccccc}
     & $g^{(1)}$ & $g^{(2)}$ & $g^{(3)}$ & $g^{(4)}$ & $g^{(5)}$ & $g^{(6)}$ \\
\hline
    \# frames & 9 & 12 & 16 & 20 & 25 & 30 \\
    \# videos & 793 & 651 & 722 & 502 & 535 & 1722 \\
    mAP(\%) & 99.8\% & 94.5\% & 93.8\% & 92.7\% & 86\% & 71.6\% 
  \end{tabular}}
\end{center}
\label{tbl:perfGatesActivityNet}
\end{table}

\begin{figure}[!ht]
\begin{center}
\begin{tabular}{cc}
\includegraphics[width=0.423\columnwidth]{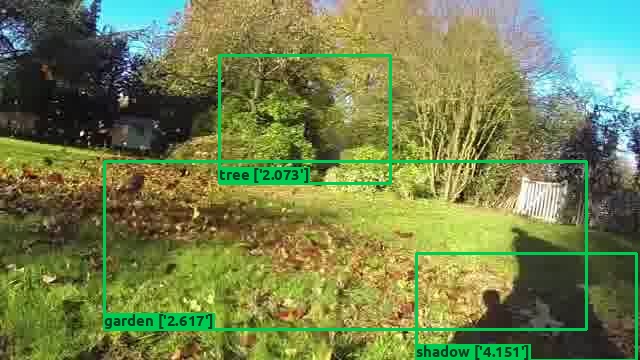} &
\includegraphics[width=0.423\columnwidth]{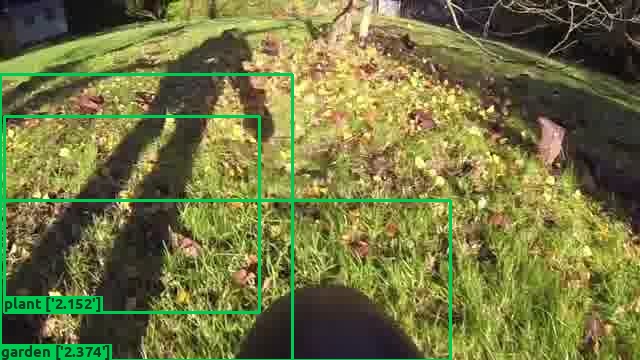} \\
\multicolumn{2}{c}{``Blowing leaves''} \\
\includegraphics[width=0.423\columnwidth]{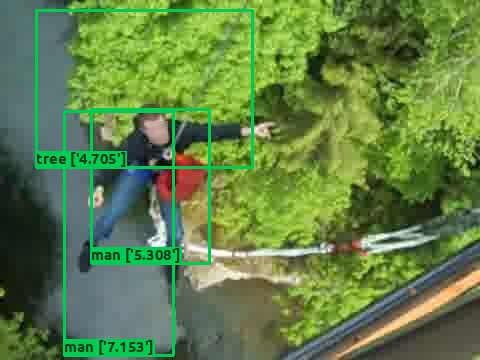}&
\includegraphics[width=0.423\columnwidth]{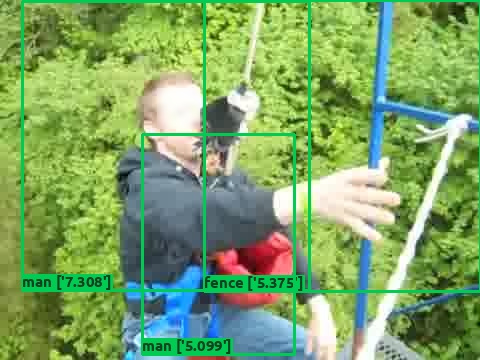} \\
\multicolumn{2}{c}{``Bungee jumping''}  \\
\includegraphics[width=0.423\columnwidth]{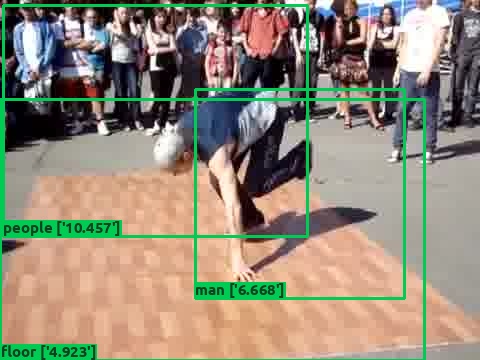}&
\includegraphics[width=0.423\columnwidth]{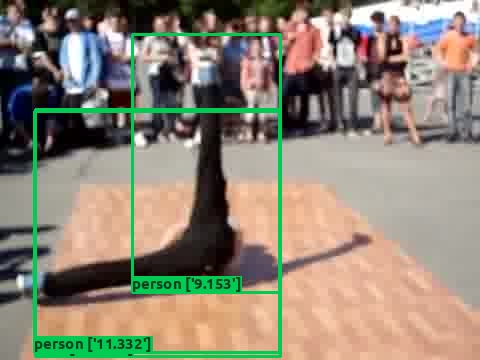} \\
\multicolumn{2}{c}{``Breakdancing''} \\
\includegraphics[width=0.423\columnwidth]{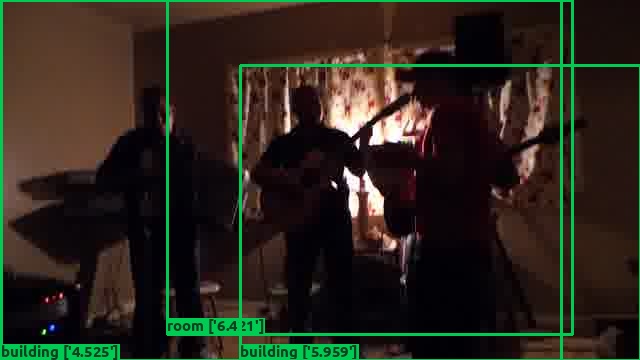}&
\includegraphics[width=0.423\columnwidth]{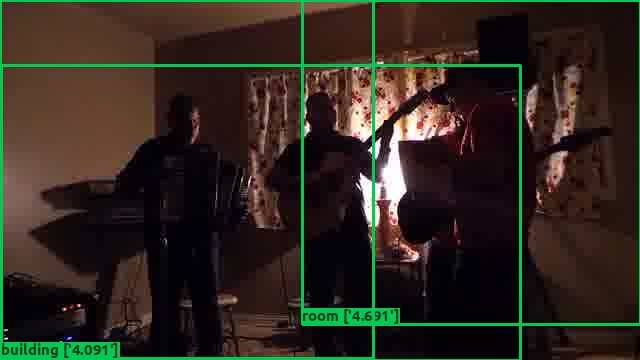} \\
\multicolumn{2}{c}{``Playing accordion'' (predicted: ``Playing guitarra'')}  \\
\includegraphics[width=0.423\columnwidth]{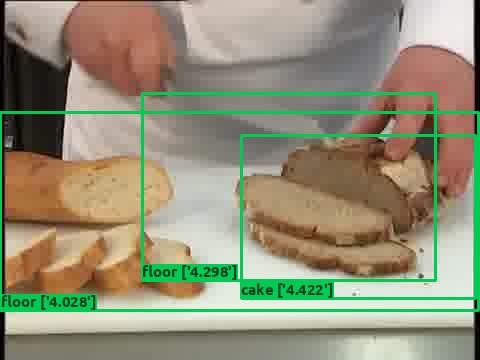}&
\includegraphics[width=0.423\columnwidth]{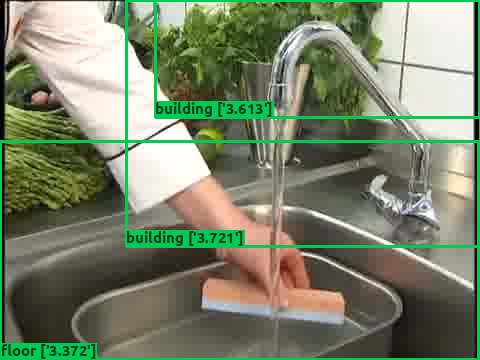} \\
\multicolumn{2}{c}{``Sharpening knives'' (predicted: ``Making a sandwich'')} \\
\includegraphics[width=0.423\columnwidth]{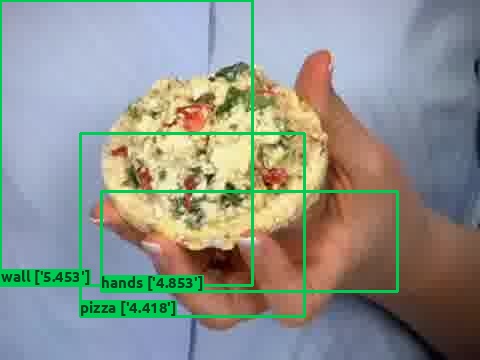} &
\includegraphics[width=0.423\columnwidth]{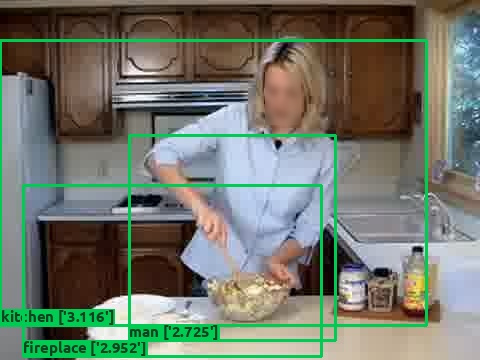} \\
\multicolumn{2}{c}{``Making a salad'' (predicted: ``Having an ice-cream'')}
\end{tabular}
\end{center}
\caption{Each row of the figure provides an explanation example produced by Gated-ViGAT for a test video from ActivityNet.
An explanation consists of the two most salient frames, and, in each frame the three most salient objects, derived using Gated-ViGAT.  
The first three examples correspond to correctly classified videos while the last three to miscategorized ones.}
\label{fig:explanationExamples}
\end{figure}

\begin{figure*}[!ht]
\begin{center}
\begin{tabular}{ccccc} 
\rotatebox[origin=t]{90}{\scriptsize WiD-based policy} &
\includegraphics[align=c,width=0.35\columnwidth]{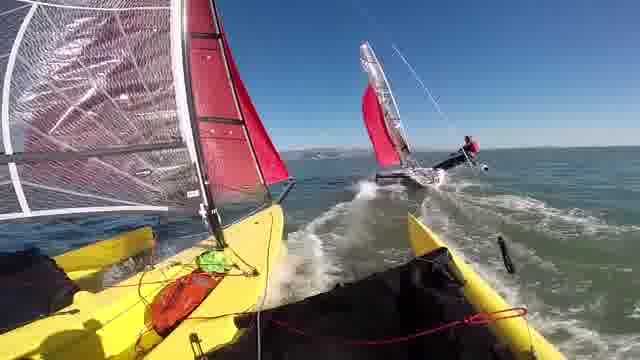} &
\includegraphics[align=c,width=0.35\columnwidth]{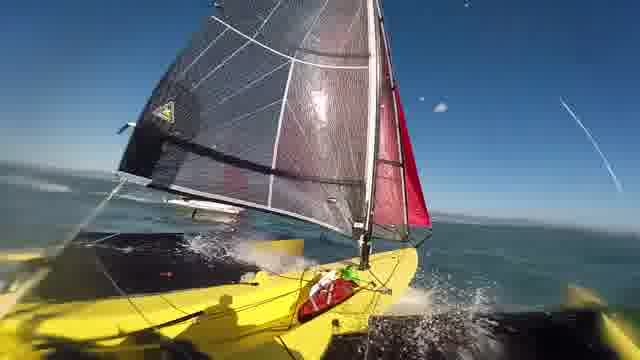} &
\includegraphics[align=c,width=0.35\columnwidth]{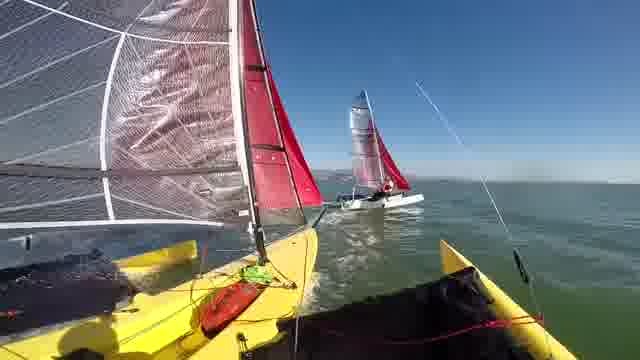} &
\includegraphics[align=c,width=0.35\columnwidth]{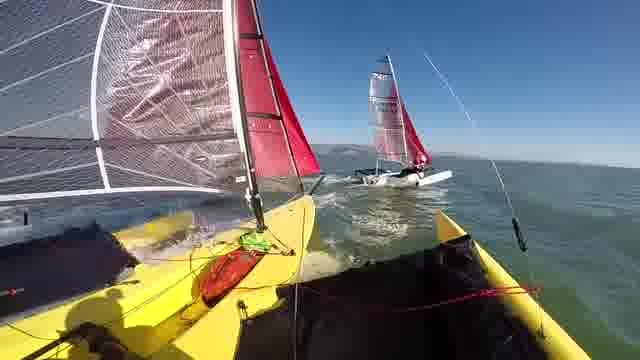} \vspace{0.08cm} \\
\rotatebox[]{90}{{\scriptsize Proposed policy}} &
\multicolumn{1}{c}{\parbox{1.5cm}{\scriptsize \textit{Same as above}}} &
\includegraphics[align=c,width=0.35\columnwidth]{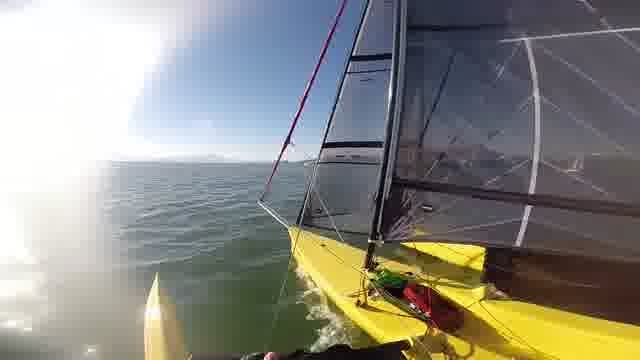} &
\includegraphics[align=c,width=0.35\columnwidth]{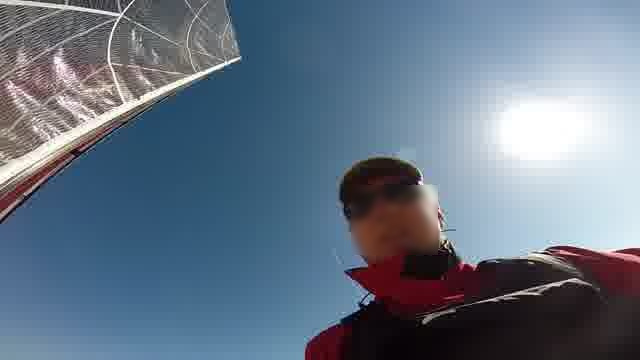} &
\includegraphics[align=c,width=0.35\columnwidth]{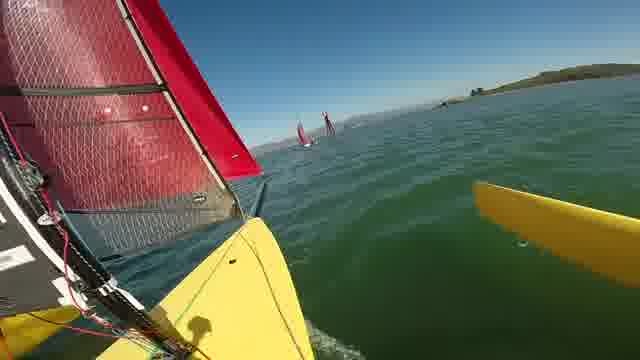} \vspace{0.08cm} \\
\multicolumn{5}{c}{``Sailing''} \\
\rotatebox[origin=t]{90}{\scriptsize WiD-based policy} &
\includegraphics[align=c,width=0.35\columnwidth]{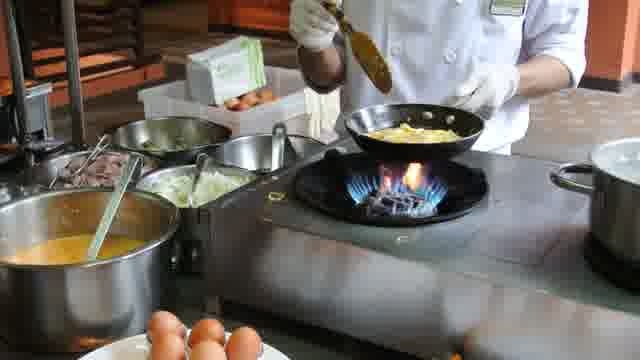} &
\includegraphics[align=c,width=0.35\columnwidth]{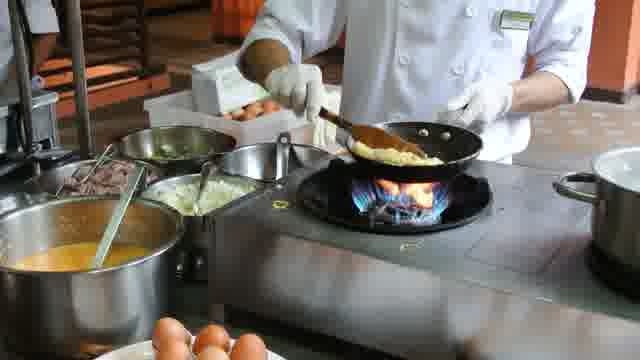} &
\includegraphics[align=c,width=0.35\columnwidth]{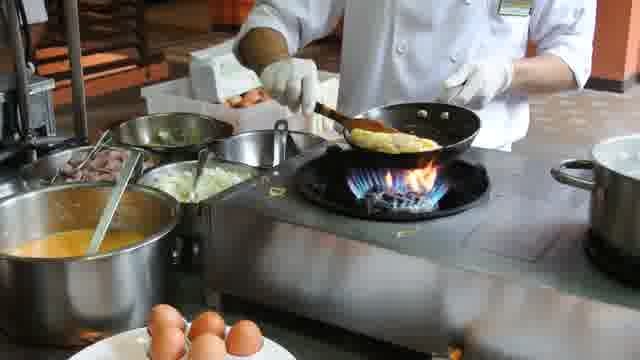} &
\includegraphics[align=c,width=0.35\columnwidth]{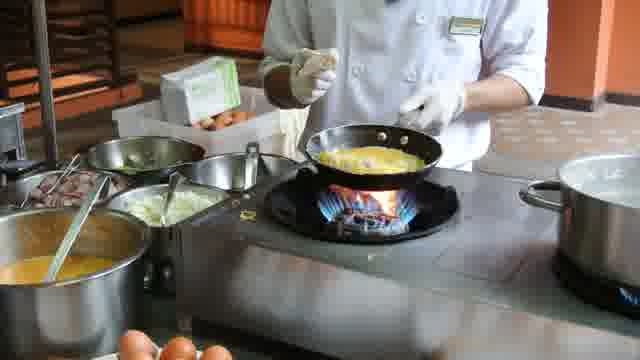} \vspace{0.08cm} \\
\rotatebox[]{90}{{\scriptsize Proposed policy}} &
\multicolumn{1}{c}{\parbox{1.5cm}{\scriptsize \textit{Same as above}}} &
\includegraphics[align=c,width=0.35\columnwidth]{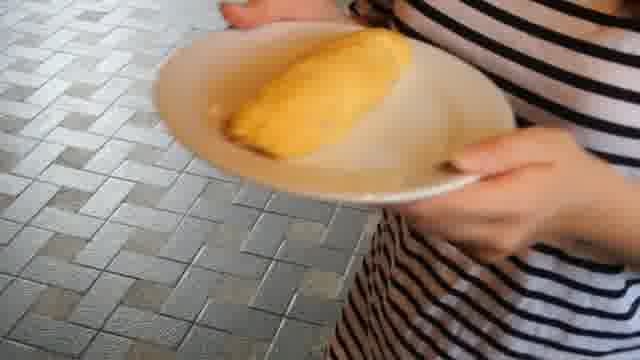} &
\includegraphics[align=c,width=0.35\columnwidth]{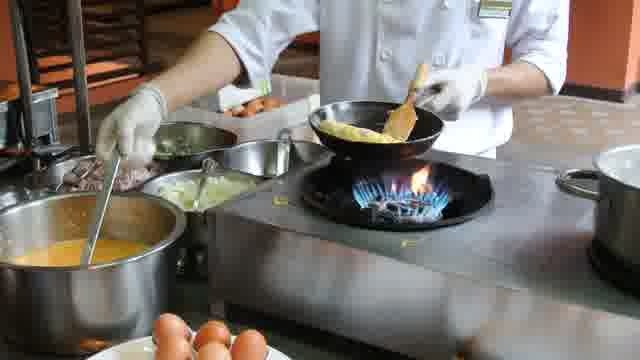} &
\includegraphics[align=c,width=0.35\columnwidth]{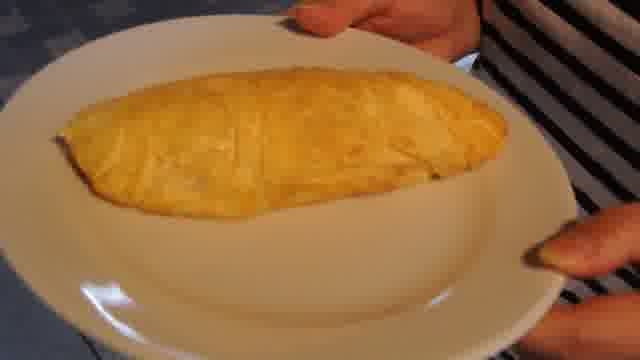} \vspace{0.08cm} \\
\multicolumn{5}{c}{``Making an omelette''}
\end{tabular}
\end{center}
\caption{Proposed vs WiD-based policy.
We see that the proposed frame selection policy can provide a broader overview of the event in the video.
For instance, in the first video example belonging to a ``Sailing'' event, the proposed approach provides frames depicting the sailor and the sailing boat from different viewing angles.
The same is observed in the second example, labeled ``Making an omelette'', where the frames selected with our approach alternate between close and distant views of the event.}
\label{fig:PolicyComparison}
\end{figure*}

\subsection{Explanations of event recognition results}
\label{ssec:ExplanEventRecognitResults}

In contrast to the majority of the top-down approaches, the proposed approach can provide explanations at object- and frame-level about the event recognition outcome.
This is done by exploiting the bottom-up information, i.e., by ranking the extracted objects and sampled frames of the test video using the corresponding WiDs, as explained in Section \ref{ss:Inference}.
For instance, explanations for six test videos of ActivityNet dataset are shown in Fig. \ref{fig:explanationExamples}.
Each row of this figure corresponds to a specific test video, where, the first three rows represent correctly classified videos, and the last three misclassified ones. 
The explanations consist of two frames and three objects per frame, as selected by Gated-ViGAT.

From the obtained results, for instance, in the correctly recognized ``Blowing leaves'' video (first row), we can see the ``person'', ``garden'' and ``tree'' objects, which are important cues for categorizing this video to the said event;
and similarly, the same is true for the objects ``man'', ``tree'', ``fences'', and, ``person'', ``floor'', for the correctly classified videos ``Bungee jumping'' (second row)  and ``Breakdancing'' (third row), respectively.
In the video ``Playing accordion'', we see that the focus is on the middle person playing guitar, misleading Gated-ViGAT to incorrectly categorize this video as ``Playing guitarra''.
Finally, regarding the videos ``Sharpening knives'' and ``Making a salad'' miscategorized as ``Making a sandwich'' and ``Having an ice-cream'', the most salient frames depict a person cutting bread and holding an object resembling an ice-cream, respectively, explaining why these videos have been misclassified.
We should also note that although in some cases the object annotations are inaccurate, which is attributed to object detector imperfections, the focus of Gated-ViGAT (i.e. the regions covered by the bounding boxes of the objects) is on the area where the underlying event is actually occurring, helping the user of the model to understand why the video was categorized in the said event class.

\subsection{Ablation study on the frame selection policy}
\label{ssec:AblationStudy}

The following frame selection policies are evaluated.
\textbf{a)} Random: $\Theta$ frames are randomly selected for deriving both the global (\ref{E:globalVidFea}) and local (\ref{E:localFrameFea}) feature vectors.
\textbf{b)}  WiD-based: $\Theta$ frames are used, as in (a), but are selected according to their global WiD values (\ref{E:globalVidFea}).
\textbf{c)}  Random local: all $P$ frames are used to derive global feature vectors (\ref{E:globalVidFea}); then, $\Theta$ frames are randomly selected for deriving the local feature vectors (\ref{E:localFrameFea}).
\textbf{d)}  WiD-based local: as above, all $P$ frames are used to derive global feature vectors (\ref{E:globalVidFea}); then, $\Theta$ frames are selected according to their global WiD values (\ref{E:globalVidFea}), and, these frames are used for computing the local feature vectors (\ref{E:localFrameFea}).
\textbf{e)}  FrameExit policy: $\Theta$ frames are selected using the frame selection policy described in \cite{GhodratiCVPR2021}.
\textbf{f)}  Proposed policy: all $P$ frames are used to derive global feature vectors (\ref{E:globalVidFea}); then, $\Theta$ frames are selected using the proposed frame selection policy (Section \ref{sss:Policy}) for deriving the local features.
\textbf{g)}  Gated-ViGAT (proposed): here, in addition to the proposed frame selection policy, as in (f), the gating mechanism is also used for selecting $\Theta$ frames per video on average, for deriving local features.

All policies are evaluated using the mAP(\%) performance measure for $\Theta = 10, 20, 30$ frames on the ActivityNet.
From the obtained results, shown in Table \ref{tbl:policyAblation}, we observe that the Gated-ViGAT policy provides the best performance.
We assume that this is because the proposed policy selects diverse frames (due to the exploitation of frame dissimilarity scores (\ref{E:frameDissimilarity}), (\ref{E:widUpdate})) and at the same time with high explanation power (due to the WiD-based values (\ref{E:policyFirstFrame}), (\ref{E:widUpdate})), thus representing well the overall video content.
It is also interesting to see that even without utilizing the gating component of Gated-ViGAT  our proposed policy (f) achieves the second-best performance, surpassing the FrameExit policy that is the best policy reported in the literature.

\begin{table}[!ht]
\caption{
Comparison of different frame selection policies in terms of recognition performance (mAP(\%)) on ActivityNet.}
\begin{center}
{%\footnotesize
\begin{tabular}{l|ccc}
      policy / \# frames & 10 & 20 & 30 \\
\hline
   Random & 83\%	& 85.5\% & 86.5\% \\
   WiD-based & 84.9\% & 86.1\%	& 86.9\% \\
   Random on local  & 85.4\% & 86.6\%	& 86.9\% \\
   WiD-based on local & 86.6\% & 87.1\% & 87.5\% \\
   FrameExit  policy \cite{GhodratiCVPR2021} & 86.2\% & \textit{87.3\%} & 87.5\% \\
   Proposed policy & \textit{86.7\%} & \textit{87.3\%} & \textit{87.6\%} \\
    Gated-ViGAT (proposed) & \textbf{86.8\%} & \textbf{87.5\%} & \textbf{87.7\%} \\ 
  \end{tabular}}
\end{center}
\label{tbl:policyAblation}
\end{table}

To gain further insight into the proposed frame selection policy (f) (Section \ref{sss:Policy}) we provide a visual comparison of it with the WiD-based one (b) in Fig. \ref{fig:PolicyComparison}.
Specifically, the four most salient frames selected by each policy for two different videos of ActivityNet are shown at each row of the figure.
We observe that the frames selected using the WiD-based policy (first and third row) are quite alike, e.g. see the ``Sailing'' video example where all frames depict the sailing boat from similar viewing angles; and the ``Making an omelette'' example where all frames show almost the same cooking scene (the cook, frying pan, etc.) in distant view. 
In contrast, the proposed policy (second and fourth row) selects frames very relevant to the event but also dissimilar to each other, thus obtaining a broader view of the video event.
For instance, see the ``Sailing'' video example, where the selected frames depict the sailor and the sailing boat from quite different viewing angles; and, similarly the ``Making an omelette'' example, where two frames depict the cook with the frying pan preparing the omelette in distant view, and the other two show in close view the prepared omelette in the plate.

\section{Conclusion}
\label{sec:Conclusions}

We presented Gated-ViGAT, an efficient bottom-up approach for event recognition in video.
Specifically, a new policy algorithm for selecting the most salient and diverse frames of the video and a gating component combining convolutional layers and a graph attention network to learn more effectively the long-term dependencies of the video were proposed.
The evaluation of Gated-ViGAT on two popular video datasets (ActivityNet, MiniKinetics) showed the efficacy of the method in terms of  both recognition performance and computational complexity.
Possible future work directions include the investigation of faster object detectors and ViT backbones \cite{YixingIEEEWFIoT2020, ZhangNips2021} and the extension of Gated-ViGAT for online event recognition in streaming applications \cite{Wu_2021_CVPR}.

%\bibliographystyle{IEEEtran}
%\bibliography{IEEEabrv,main}


\end{document}